\documentclass{article}
\usepackage{spconf,amsmath,graphicx,hyperref}
\usepackage{booktabs}
\usepackage{amsmath}
\usepackage{float}   
\usepackage{IEEEtrantools}
\usepackage{caption}
\usepackage{amssymb}
\usepackage{microtype}

\def\L{{\cal L}}

\title{Joint-OCTAMamba: An OCTA  Joint Segmentation network based on Feature enhanced Mamba }
\twoauthors
 {A. Chuang Liu}
 {Northeastern University\\
 Department of Computer Science
 \\and Engineering\\
 Shenyang, China}
 {B. Nan Guo}
 {Northeastern University\\
 Department of Computer Science\\
 and Engineering\\
 Shenyang, China}

\begin{document}
\ninept
\maketitle
\begin{abstract}
OCTA is a crucial non-invasive imaging technique for diagnosing and monitoring retinal diseases like diabetic retinopathy, age-related macular degeneration, and glaucoma. Current 2D-based methods for retinal vessel (RV) segmentation offer insufficient accuracy. To address this, we propose RVMamba, a novel architecture integrating multiple feature extraction modules with the Mamba state-space model. Moreover, existing joint segmentation models for OCTA data exhibit performance imbalance between different tasks. To simultaneously improve the segmentation of the foveal avascular zone (FAZ) and mitigate this imbalance, we introduce FAZMamba and a unified Joint-OCTAMamba framework. Experimental results on the OCTA-500 dataset demonstrate that Joint-OCTAMamba outperforms existing models across evaluation metrics.The code is available at https://github.com/lc-sfis/Joint-OCTAMamba.
\end{abstract}
\section{Introduction}
\label{sec:intro}
Optical Coherence Tomography Angiography (OCTA) is a key non-invasive imaging modality for visualizing the retinal microvasculature, crucial for diagnosing diseases like diabetic retinopathy and glaucoma~\cite{Spaide2018OCTA}\cite{kashani2017optical}. However, accurately co-segmenting RV and FAZ remains a significant challenge due to complex vessel structures, imaging noise, and other artifacts.

The recently proposed Mamba architecture~\cite{gu2023mamba} presents a highly efficient alternative to Transformers\cite{vaswani2017attention} by replacing their quadratic $O(L^2)$ complexity with a linear $O(L)$ selective state space model. The pioneering OCTA-Mamba~\cite{zou2025octamamba} has already demonstrated its effectiveness for vessel segmentation. However, the model's RV segmentation accuracy has not yet reached the state-of-the-art level. Meanwhile, current joint segmentation models are hampered by one of two issues: unbalanced performance between tasks\cite{masset2000fargo,hu2022joint}, or high training complexity\cite{jiang2025joint}.

To solve these problems, we first developed two specialized architectures: RVMamba and FAZMamba. These are then combined to form the comprehensive Joint-OCTAMamba model. The main contributions are:

$(1)$ Task-Specific Mamba Architectures: We propose two specialized models: RVMamba, featuring a Hybrid Directional Feature Extractor (HDFE) and a Vessel Multi-Attention Fusion (VMAF) module for precise vasculature mapping, and FAZMamba, which shares a similar architectural design with RVMamba, features the addition of a Compact FAZ Enhancement Block (CFEB) to specifically enhance FAZ delineation.

$(2)$ Synergistic Multi-Task Framework: We introduce Joint-OCTAMamba, the first framework to leverage the Mamba architecture for simultaneous, synergistic segmentation of RV and FAZ. To ensure rigorous evaluation, we also develop an isomorphic Swin-Transformer baseline, Joint-OCTAFormer.

$(3)$ Extensive Empirical Validation: Comprehensive experiments on the OCTA-500 dataset (3 mm and 6 mm scans) demonstrate that our proposed Joint-OCTAMamba achieves leading performance on RV segmentation and state-of-the-art performance on FAZ segmentation, confirming the superior performance and efficacy of our approach for OCTA analysis. 

\begin{figure*}[t]   
  \centering
  \includegraphics[width=0.95\textwidth]{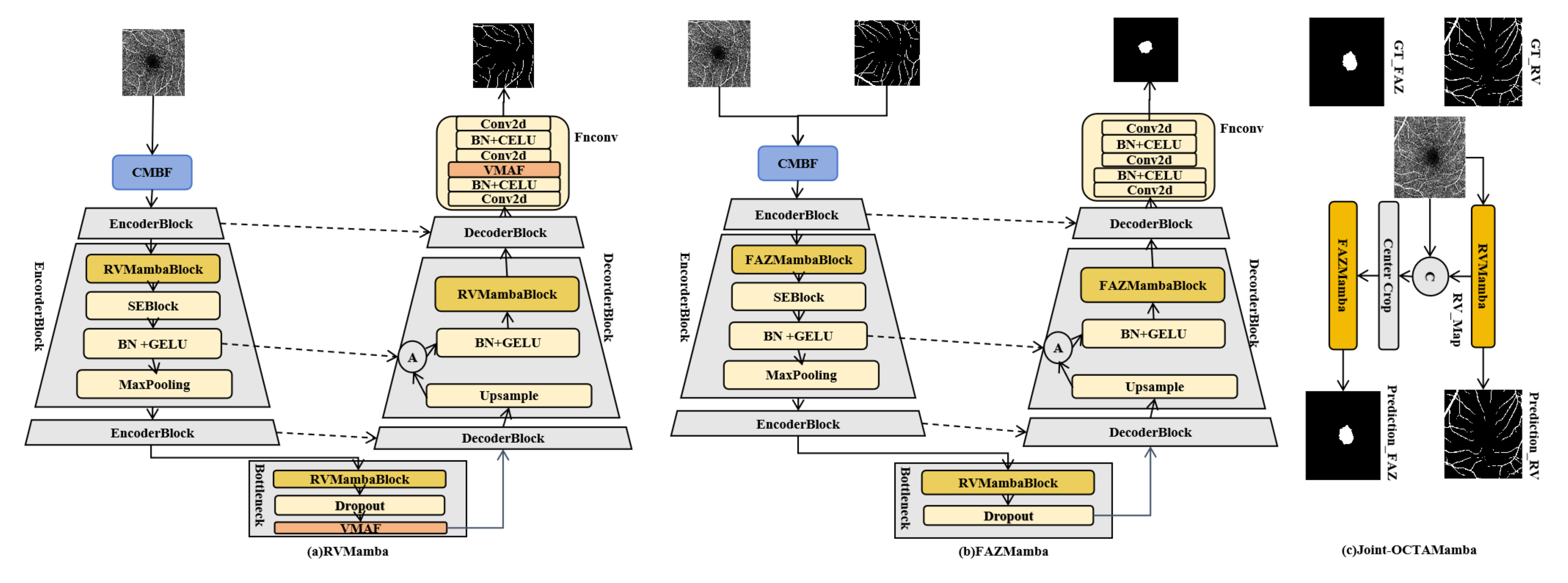}
  \caption{Overview of the proposed Joint-OCTAMamba framework. }
  \label{fig:1}
\end{figure*}

\section{Related Work}
The segmentation of retinal structures in OCTA images has progressed through several architectural paradigms, each addressing the limitations of its predecessors. Early approaches relied on CNNs like U-Net\cite{ronneberger2015u} to capture coarse vasculature but were inherently constrained by limited receptive fields. To overcome this, OCT2Former\cite{tan2023oct2former} used Transformer blocks to model long-range dependencies, but introduced quadratic computational complexity. Alternative strategies also emerged: IPN\cite{li2020image} and IPN-V2\cite{li2020ipn} used 3D–2D backbones for volumetric data but required costly 3D annotations and large GPU memory, while foundation models like MedSAM\cite{ma2024segment} and its variants SAM-OCTA\cite{wang2024sam} and SAM-OCTA2\cite{chen2025sam} delivered robust results yet remained heavyweight. Multi-task learning was developed to leverage anatomical coupling, with notable approaches including FARGO\cite{masset2000fargo} using independent branches, Joint-Seg\cite{hu2022joint} featuring shared encoders, and PL-Joint-Seg\cite{jiang2025joint} incorporating 3D training. Despite accuracy gains, these joint models ultimately inherited the weaknesses of their backbones, suffering from either convolutional limitations or 3D overhead. Addressing these persistent challenges of complexity and efficiency, the recent Mamba architecture, built on Selective State-Space Models, attains global modeling in linear complexity. The successful application of Mamba to RV segmentation by OCTAMamba\cite{zou2025octamamba} highlights its potential.Building upon this success, we identify a crucial research gap: extending this efficiency to the more complex joint segmentation of RV and FAZ.

\begin{figure*}[t]
  \includegraphics[width=0.95\textwidth]{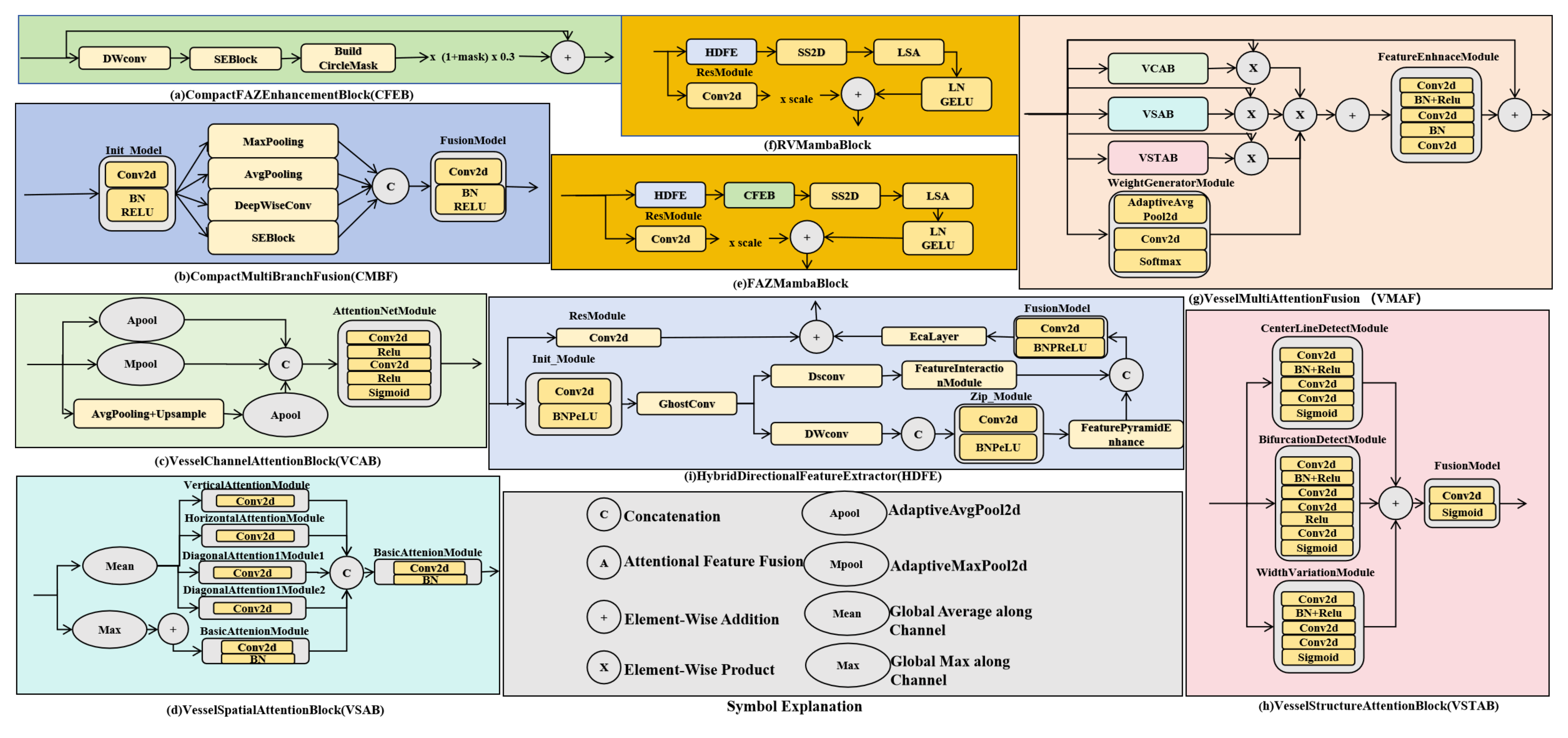}
  \caption{Details of the Joint-OCTAMamba framework.}
  \label{fig:overview}
\end{figure*}
\section{Method}
\label{sec:pagestyle}
Joint-OCTAMamba utilizes a two-stage cascaded pipeline to segment RV and FAZ, as shown in Fig~\ref{fig:1}(c). This design decouples the tasks to prevent feature interference.
In the first stage, RVMamba processes the input OCTA volume to produce  a RVMap. This map is then concatenated channel-wise with the original volume, serving as a spatial prior. In the second stage, the combined data is center-cropped(Region of Interest) and passed to the FAZMamba module to generate the final FAZ segmentation. The architecture of each module is detailed below.
\subsection{RVMamba and FAZMamba}

\begin{figure}[ht]
  \centering
  \includegraphics[width=\linewidth]{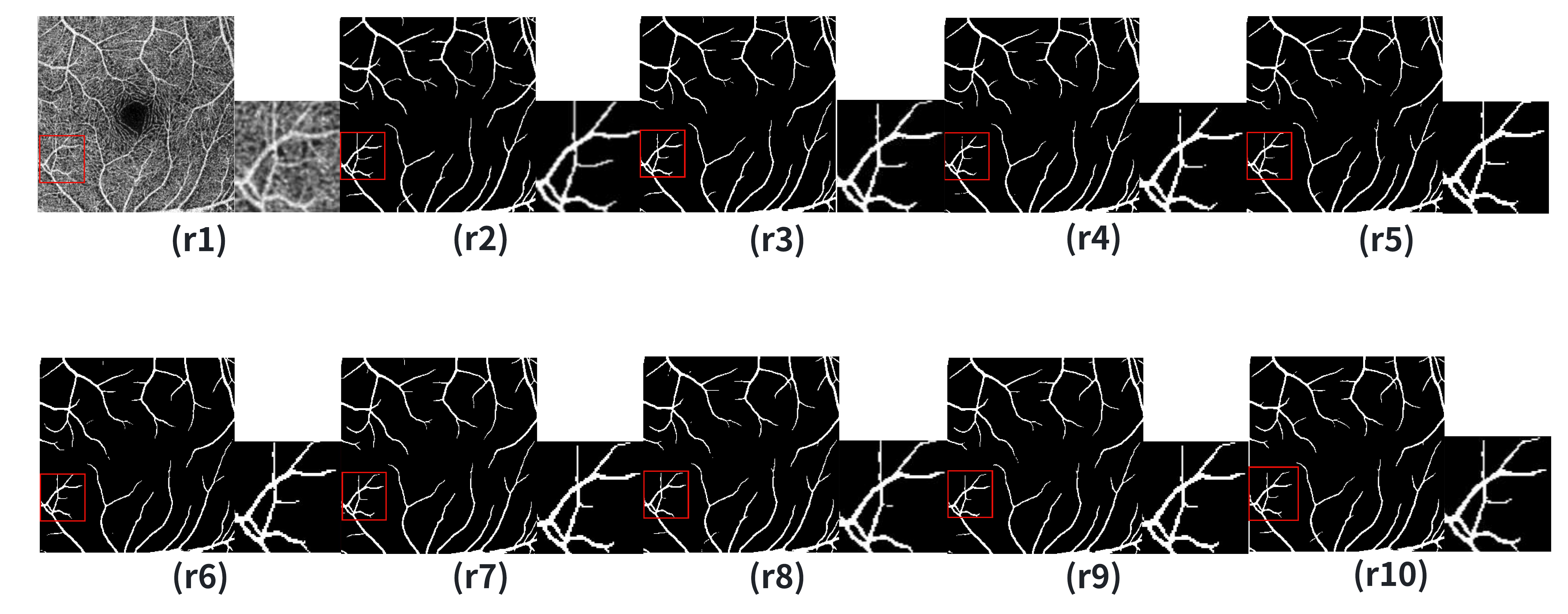}
 \caption{RV ablation visualizations at higher magnification. (r1) Raw OCTA  No.10466; (r2) GT; (r3) Baseline; (r4-r10) The baseline with progressive additions of the CMBF (C), HDFE (H), and VMAF (V) modules. The configurations are: (r4) C; (r5) H; (r6) V; (r7) C+V; (r8) C+H; (r9) V+H; (r10) C+H+V.}
  \label{fig:rv_abla}
\end{figure}

\begin{figure}[ht]
  \centering
  \includegraphics[width=\linewidth]{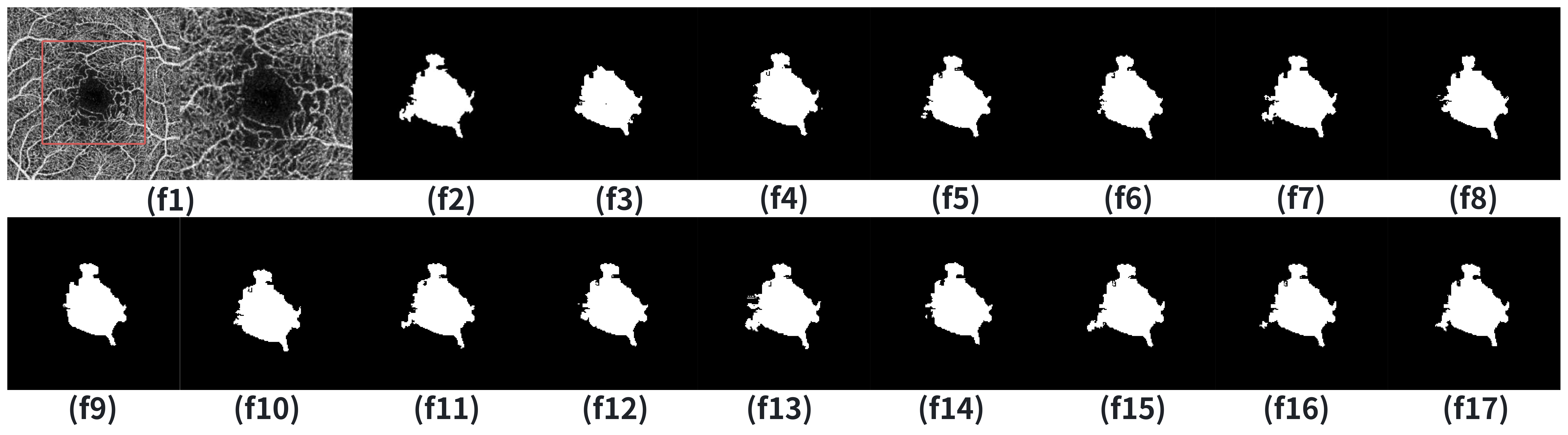}
\caption{FAZ ablation visualizations at higher magnification. (\textit{BL}: Baseline; \textit{+} denotes module addition).
(f1) Raw OCTA No.10496; (f2) GT; (f3) BL;
(f4-f17) The baseline with progressive additions of the following modules: CMBF (C), HDFE (H), VMAF (V), CFEB (F), ROI (O), and RV (R). The configurations are:
(f4) C; (f5) H; (f6) V;
(f7) C+H; (f8) C+V; (f9) H+V;
(f10) C+H+V; (f11) C+H+F; (f12) C+H+O; (f13) C+H+R;
(f14) C+H+F+O; (f15) C+H+R+O; (f16) C+H+R+F;
(f17) C+H+R+F+O.}
  \label{fig:faz_abla}
\end{figure}

The RVMamba adopts a U-Net skeleton,as show in Fig\ref{fig:1}(a). A Compact Multi-Branch Fusion (CFMB) pre-processes the image to align channels, followed by three encoder blocks that down-sample while enriching vessel cues. A bottleneck of one RVMamba block, Dropout and Vessel Multi-Attention Fusion (VMAF) injects global context and vessel priors. Three decoder blocks up-sample with  Attentional Feature Fusion\cite{dai2021attentional}, then the final convolution module (Fnconv) yields the vessel probability map. The FAZMamba reuses the same encoder–decoder architecture, while drops VMAF module and takes CFEB  due to ablation experiment (as depicted in Table\ref{fig:rv_abla} and Table\ref{fig:faz_abla}). 

\subsection{Compact Multi-Branch Fusion (CMBF)}
CFMB first expands the OCTA images to 32 channels, splits it into four 8-channel branches that apply maxpool, avgpool,Depthwise Separable Convolution, and SE recalibration\cite{hu2018squeeze}, concatenates the results, and fuses them back to the target channel count with a 1×1 convolution,as shown in Fig\ref{fig:overview}(b).

\subsection{Vessel Multi-Attention Fusion (VMAF)}
The VMAF module, illustrated in Fig\ref{fig:overview}(g), adaptively combines three specialized attention mechanisms-channel, spatial, and structural-to refine vascular feature representation. Each component is specifically designed to capture unique properties of vessel morphology, and their outputs are integrated via learnable weights,then the three attention maps are adaptively fused, enhanced by the EnhanceModule, and added to the residual connection.

\textbf{Vessel Channel Attention Block (VCAB)}, as shown in Fig\ref{fig:overview}(c). Global average and max pooling capture holistic statistics; local average pooling followed by up-sampling and re-pooling captures local-global statistics. These three descriptors are concatenated and processed by a $1 \times 1$ convolution–ReLU–depthwise separable convolution–$1 \times 1$ convolution stack, followed by Sigmoid to generate channel-wise attention weights in $[0,1]$.

\textbf{Vessel Spatial Attention Block (VSAB)}, as shown in Fig\ref{fig:overview}(d). The input is first pooled (average and max) and concatenated to obtain a base attention map. Directional attention is extracted via horizontal and vertical convolutions, while diagonal attention is captured by rotating the pooled feature, convolving, and rotating back. All three maps are concatenated, reduced by a $1 \times 1$ convolution, and activated by Sigmoid to yield spatial attention.

\textbf{Vessel Structure Attention Block (VSTAB)}, as shown in Fig\ref{fig:overview}(h).Center-line, bifurcation, and width-variation attention maps are generated via dedicated convolutions, dilated convolutions, and Sigmoid activations. These maps are concatenated and fused by a $1 \times 1$ convolution to produce the final structural attention.

\subsection{Hybrid Directional Feature Extractor (HDFE)}
As shown in Fig~\ref{fig:overview}(i), HDFE serves as the primary feature extractor within our architecture. 
Unlike the feature extraction strategy in OCTAMamba\cite{zou2025octamamba}, which relies exclusively on dilated convolutions and an ECA\cite{wang2020eca} layer, HDFE integrates a comprehensive multi-branch design. 
Given an input tensor $\mathbf{X}\!\in\!\mathbb{R}^{H\times W\times C}$, it first generates foundational features using a 1$\times$1 convolution with PReLU\cite{he2015delving} and a GhostModule\cite{han2020ghostnet}. 
These features are then processed in two parallel streams: (1) a direction-aware stream that uses dual $1\!\times\!7$ and $7\!\times\!1$ Dynamic Snake Convolutions (DSConv)\cite{qi2023dynamic} to encode orientation cues, refined by a feature interaction module; and (2) a multi-scale stream that captures context using dilated convolutions with rates of $\{2,4,6\}$. 
The features from this multi-scale stream are enhanced by our custom Feature Pyramid Enhance module, which is inspired by the principles of FPN\cite{lin2017feature}. 
Finally, the outputs of both streams are adaptively fused, re-weighted by an ECA layer, and added to a residual connection to produce the final features.
\subsection{RVMambaBlock and FAZMambaBlock}
RVMambaBlock, shown in Fig\ref{fig:overview}(f), the HDFE block is followed by V-SS2D. In V-SS2D, the input is separated into a main path and a gate path. The main path undergoes 2-D depthwise convolution and Selective Scan for global spatial-state modeling, while the gate path is processed by VMAF to generate vessel-aware attention weights. The two paths are then fused using a SiLU\cite{elfwing2018sigmoid} gating mechanism and element-wise multiplication to emphasize fine vessels.For training stability, a residual connection is employed. The output is then passed through an LSA\cite{cheng2025mamba} module to enhance generalization capabilities. Finally, it is normalized, subjected to a non-linear activation, and a final residual connection is added.

FAZMambaBlock follows the same pipeline but drops VMAF and takes a \textbf{Compact FAZ Enhancement Block (CFEB)}, Fig\ref{fig:overview}(a). CFEB first applies a $7 \times 7$ grouped convolution with dilation $= 2$ to capture annular FAZ context. An SE module\cite{hu2018squeeze} re-calibrates channels, and an optional circular mask centered at $(H/2, W/2)$ with radius $\min(H,W)/4$ enlarges the foveal region. The refined feature is fused with the original input via a residual scale of $0.3$, emphasizing FAZ boundaries.
\begin{table*}[ht]
\centering
\caption{RV and FAZ Segmentation Results on OCTA-500 dataset (Mean$\pm$STD)}
\label{tab:segmentation_results}
\resizebox{\textwidth}{!}{
\begin{tabular}{lcccccccc}
\toprule
\textbf{Method } & \multicolumn{2}{c}{\textbf{RV (3M)}} & \multicolumn{2}{c}{\textbf{RV (6M)}} & \multicolumn{2}{c}{\textbf{FAZ (3M)}} & \multicolumn{2}{c}{\textbf{FAZ (6M)}} \\
\cmidrule(lr){2-3}\cmidrule(lr){4-5}\cmidrule(lr){6-7}\cmidrule(lr){8-9}
& Dice $\uparrow$ & Jaccard $\uparrow$ & Dice $\uparrow$ & Jaccard $\uparrow$ & Dice $\uparrow$ & Jaccard $\uparrow$ & Dice $\uparrow$ & Jaccard $\uparrow$ \\
\midrule
U-Net \cite{ronneberger2015u} & 0.9068\small{$\pm$0.0205} & 0.8301\small{$\pm$0.0329} & 0.8876\small{$\pm$0.0253} & 0.7987\small{$\pm$0.0392} & 0.9747\small{$\pm$0.0130} & 0.9585\small{$\pm$0.0244} & 0.8770\small{$\pm$0.1779} & 0.8124\small{$\pm$0.2018} \\
IPN $^\dag$\cite{li2020image} & 0.9062\small{$\pm$0.0208} & 0.8325\small{$\pm$0.0778} & 0.8864\small{$\pm$0.0321} & 0.7973\small{$\pm$0.0492} & 0.9505\small{$\pm$0.0479} & 0.9091\small{$\pm$0.0798} & 0.8802\small{$\pm$0.0991} & 0.7980\small{$\pm$0.1381} \\
IPN V2+$^\dag$ \cite{li2020ipn} & \textbf{0.9274\small{$\pm$0.0208}} & \textbf{0.8667\small{$\pm$0.0588}} & 0.8941\small{$\pm$0.0274} & 0.8095\small{$\pm$0.0432} & 0.9755\small{$\pm$0.0238} & 0.9532\small{$\pm$0.0419} & 0.9084\small{$\pm$0.0893} & 0.8423\small{$\pm$0.1269} \\
FARGO \cite{masset2000fargo} & 0.9168\small{$\pm$0.0205} & 0.8470\small{$\pm$0.0334} & 0.8915\small{$\pm$0.0239} & 0.8050\small{$\pm$0.0375} & 0.9839\small{$\pm$0.0092} & 0.9684\small{$\pm$0.0176} & 0.9272\small{$\pm$0.0674} & 0.8701\small{$\pm$0.1069} \\
Joint-Seg \cite{hu2022joint} & 0.9113\small{$\pm$0.0209} & 0.8378\small{$\pm$0.0340} & \underline{0.8972\small{$\pm$0.0185}} & \underline{0.8117\small{$\pm$0.0311}} & 0.9843\small{$\pm$0.0089} & 0.9693\small{$\pm$0.0170} & 0.9051\small{$\pm$0.1113} & 0.8424\small{$\pm$0.1544} \\
OCT2Former \cite{tan2023oct2former} & 0.9193\small{$\pm$0.0206} & 0.8513\small{$\pm$0.0339} & 0.8945\small{$\pm$0.0215} & 0.8099\small{$\pm$0.0177} & -- & -- & -- & -- \\
SAM-OCTA \cite{wang2024sam} & 0.9199\small{$\pm$N/A} & 0.8520\small{$\pm$N/A} & 0.8869\small{$\pm$N/A} & 0.7975\small{$\pm$N/A} & 0.9838\small{$\pm$N/A} & 0.9692\small{$\pm$N/A} & 0.9073\small{$\pm$N/A} & 0.8473\small{$\pm$N/A} \\
OCTA-Mamba \cite{zou2025octamamba} & 0.8450\small{$\pm$N/A} & 0.7323\small{$\pm$N/A} & 0.8231\small{$\pm$N/A} & 0.7003\small{$\pm$N/A} & -- & -- & -- & -- \\
SAM-OCTA2 \cite{chen2025sam} & 0.9207\small{$\pm$N/A} & 0.8428\small{$\pm$N/A} & 0.8923\small{$\pm$N/A} & 0.8046\small{$\pm$N/A} & 0.9833\small{$\pm$N/A} & 0.9687\small{$\pm$N/A} & \underline{0.9284\small{$\pm$N/A}} & 0.8733\small{$\pm$N/A} \\
PL-Joint-Seg$^\dag$ \cite{jiang2025joint} & 0.9165\small{$\pm$0.0194} & 0.8498\small{$\pm$0.0298} & \textbf{0.8982\small{$\pm$0.0162}} & \textbf{0.8201\small{$\pm$0.0288}} & \underline{0.9848\small{$\pm$0.0081}} & \underline{0.9701\small{$\pm$0.0165}} & 0.9110\small{$\pm$0.0623} & 0.8522\small{$\pm$0.1010} \\
\midrule
RVMamba (ours) & 0.9210\small{$\pm$0.0015} & 0.8541\small{$\pm$0.0017} & 0.8940\small{$\pm$0.0021} & 0.8092\small{$\pm$0.0025} & 0.9812\small{$\pm$0.0031} & 0.9645\small{$\pm$0.0042} & 0.9105\small{$\pm$0.0055} & 0.8615\small{$\pm$0.0060} \\
RVFormer (ours) & 0.9200\small{$\pm$0.0010} & 0.8520\small{$\pm$0.0024} & 0.8936\small{$\pm$0.0021} & 0.8084\small{$\pm$0.0031} & 0.9783\small{$\pm$0.0053} & 0.9578\small{$\pm$0.0086} & 0.9095\small{$\pm$0.0035} & 0.8604\small{$\pm$0.0042} \\
Joint-OCTAFormer (ours) & 0.9200\small{$\pm$0.0019} & 0.8520\small{$\pm$0.0022} & 0.8936\small{$\pm$0.0029} & 0.8084\small{$\pm$0.0040} & 0.9840\small{$\pm$0.0022} & 0.9698\small{$\pm$0.0028} & 0.9267\small{$\pm$0.0041} & \underline{0.8799\small{$\pm$0.0054}} \\
Joint-OCTAMamba (ours) & \underline{0.9210\small{$\pm$0.0022}} & \underline{0.8541\small{$\pm$0.0026}} & 0.8940\small{$\pm$0.0033} & 0.8092\small{$\pm$0.0053} & \textbf{0.9849\small{$\pm$0.0024}} & \textbf{0.9705\small{$\pm$0.0039}} & \textbf{0.9330\small{$\pm$0.0049}} & \textbf{0.8871\small{$\pm$0.0075}} \\
\bottomrule
\end{tabular}
}
\begin{flushleft}
\footnotesize
* For each column, the top two best entries are highlighted in \textbf{bold} and \underline{underlined}, respectively.
``$^\dag$'' indicates that the inputs of the method include 3D OCTA volume. ``N/A'' denotes values not available.
\end{flushleft}
\end{table*}
\section{Experniment}
\label{sec:typestyle}
\subsection{Dataset and Implementation Details}
All experiments are conducted on the OCTA-500\cite{li2024octa} dataset, using its official 3M and 6M splits. Models are trained for 100 epochs on a single NVIDIA V100-32 GB GPU with a batch size of 2.The input images were maintained their original dimensions,using center-cropped(ROI) $224 \times 224$ sizes for FAZ.The optimizer employed is AdamW, and the learning rate schedule follows a OneCycleLR strategy.The maximum learning rate of $1 \times 10^{-3}$, an initial learning rate of $5 \times 10^{-4}$, and a 10-epoch warm-up, followed by cosine decay. Due to different task convergence time, three separate checkpoints are saved based on the best RV-Dice, FAZ-Dice, and their average for final testing. To prevent early overfitting, during the first 70\% of epochs, we only perform training and do not evaluate on the validation or test sets. For the final 30\% of epochs, we save models based on the validation set's DICE scores (RV, FAZ, and their average) and save checkpoints every five epochs. Reproducibility is ensured by fixing the random seed to 0. For data augmentation, we use Albumentations\cite{buslaev2020albumentations} ($p=0.2$) to apply random brightness and contrast shifts, CLAHE, a $\pm15^{\circ}$ rotation, horizontal and vertical flips, and a piecewise affine transformation to both the input image and its corresponding RV and FAZ masks. Test-Time Augmentation (TTA) is also applied by averaging predictions from the original, horizontally flipped, and vertically flipped inputs. The RV branch loss is a combination of Dice, boundary, Tversky, and Hausdorff losses: $0.6\mathcal{L}_{\text{dice}}+0.2\mathcal{L}_{\text{boundary}}+0.1\mathcal{L}_{\text{Tversky}}+0.1\mathcal{L}_{\text{Hausdorff}}$. The FAZ branch loss is a weighted sum of Dice and boundary losses: $0.8\mathcal{L}_{\text{dice}}+0.2\mathcal{L}_{\text{boundary}}$. The total loss is calculated as $\mathcal{L}_{\text{total}}=\lambda_{\text{RV}}\mathcal{L}_{\text{RV}}+\lambda_{\text{FAZ}}\mathcal{L}_{\text{FAZ}}$, where a single optimizer updates the entire network. Loss weights are set to $\lambda_{\text{RV}}=1$ and $\lambda_{\text{FAZ}}=6.1$ for 3M, and $\lambda_{\text{RV}}=1$ and $\lambda_{\text{FAZ}}=4$ for 6M to balance task complexity. Performance is evaluated using the Dice and Jaccard metrics.
\begin{align}
\mathrm{Dice} =\frac{2 \times |Y \cap P|}{|Y| + |P|} 
\end{align}
\begin{align}
\mathrm{Jaccard} =\frac{|Y \cap P|}{|Y \cup P|}
\end{align}
where $Y$ is the ground truth and $P$ is the prediction.
\subsection{Comparison With State-of-the-Art Methods}
As shown in Table~\ref{tab:segmentation_results}, Joint-OCTAMamba outperforms other existing OCTA segmentation models, achieving excellent performance on RV and FAZ segmentation while maintaining high stability. A key advantage of Joint-OCTAMamba is its efficiency, requiring only 13.22 M trainable parameters. This is significantly fewer than U-Net\cite{ronneberger2015u} (118.42 M), FARGO\cite{masset2000fargo} (328.14 M), Joint-Seg\cite{hu2022joint} (154.16 M), IPN-V2\cite{li2020ipn} (195.64 M), and our own Joint-OCTAFormer (22.88 M).
\subsection{Ablation Experiments}
To validate the effectiveness of our proposed modules and the necessity of the joint-segmentation framework, we conducted a comprehensive ablation study on OCTA-500 3M. Module-wise ablations are reported in Table\ref{tab:ablation_3m_single}, whereas the contributions of prior knowledge and region-of-interest strategies are summarized in Table\ref{tab:ablation_6m}. From Table\ref{tab:ablation_3m_single}, we observe that HDFE, CMBF, and VMAF each yield desirable improvements for RV and FAZ segmentation. However, combinations involving VMAF underperform on FAZ segmentation; hence, we removed the VMAF module in FAZ-Mamba. Table\ref{tab:ablation_6m} further demonstrates that RV prior knowledge delivers a substantial boost to FAZ results, while the CFEB structure and ROI strategy also provide consistent gains, underscoring the indispensability of our joint-segmentation design.
\begin{table}
  \centering
  \caption{Ablation study of RVMamba on OCTA-500 3M dataset.}
  \label{tab:ablation_3m_single}
  \resizebox{\linewidth}{!}{%
    \begin{tabular}{lccccccc}
      \toprule
      \textbf{ID} & \textbf{HDFE} & \textbf{VMAF} & \textbf{CMBF} &
      \textbf{RV-Dice}\,$\uparrow$ & \textbf{RV-Jaccard}\,$\uparrow$ &
      \textbf{FAZ-Dice}\,$\uparrow$ & \textbf{FAZ-Jaccard}\,$\uparrow$ \\
      \midrule
      1 & -- & -- & -- & 0.9189 & 0.8524 & 0.9750 & 0.9565 \\
      2 & $\surd$ & -- & -- & 0.9203 & 0.8532 & 0.9803 & 0.9618 \\
      3 & -- & $\surd$ & -- & 0.9199 & 0.8531 & 0.9813 & 0.9639 \\
      4 & -- & -- & $\surd$ & 0.9201 & 0.8529 & 0.9810 & 0.9633 \\
      5 & $\surd$ & $\surd$ & -- & 0.9204 & 0.8533 & 0.9800 & 0.9623 \\
      6 & $\surd$ & -- & $\surd$ & 0.9204 & 0.8533 &\textbf{ 0.9819 }& \textbf{0.9650} \\
      7 & -- & $\surd$ & $\surd$ & 0.9207 & 0.8538 & 0.9805 & 0.9623 \\
      8 & $\surd$ & $\surd$ & $\surd$ & \textbf{0.9210} & \textbf{0.8540} & 0.9812 & 0.9645 \\
      \bottomrule
      
    \end{tabular}%
  }
  \begin{flushleft}
\footnotesize
* For each column, the best entry is highlighted in \textbf{bold}
\end{flushleft}
\end{table}
\begin{table}[H]
  \centering
  \caption{Ablation study of Joint-OCTAMamba on  OCTA-500 3M dataset for the FAZ task.}
  \label{tab:ablation_6m}
  \resizebox{\linewidth}{!}{%
    \begin{tabular}{lccccccc}
      \toprule
      \textbf{ID} & \textbf{CFEB} & \textbf{ROI} & \textbf{RV-knowledge} &
      \textbf{FAZ-Dice}\,$\uparrow$ & \textbf{FAZ-Jaccard}\,$\uparrow$ \\
      \midrule
      1 & -- & -- & -- & 0.9819 & 0.9650 \\
      2 & $\surd$ & -- & -- & 0.9825 & 0.9661 \\
      3 & -- & $\surd$ & -- & 0.9823 & 0.9660 \\
      4 & -- & -- & $\surd$ & 0.9840 & 0.9701 \\
      5 & $\surd$ & $\surd$ & -- & 0.9828 & 0.9662 \\
      6 & $\surd$ & -- & $\surd$ & 0.9841 & 0.9700 \\
      7 & -- & $\surd$ & $\surd$ & 0.9847 & 0.9697 \\
      8 & $\surd$ & $\surd$ & $\surd$ & \textbf{0.9849} & \textbf{0.9706} \\
      \bottomrule
    \end{tabular}%
  }
\end{table}
\section{Conclusion}
We present Joint-OCTAMamba, a new framework for segmenting both RV and FAZ in OCTA images. By combining four custom modules with the Mamba architecture, our method greatly improves RV segmentation. At the same time, a special decoupled design for the FAZ task achieves state-of-the-art accuracy and prevents the two tasks from interfering. Mamba's efficiency allows our model to be both powerful and practical for real-world use. Tested on the OCTA-500 dataset, Joint-OCTAMamba shows superior performance, setting a new benchmark for accuracy and efficiency in its field.

\vfill\pagebreak




\bibliographystyle{IEEEbib}

\end{document}